\documentclass{article}
\usepackage{ijcai26}

\usepackage{adjustbox}
\usepackage{times}
\usepackage{soul}
\usepackage{url}
\usepackage[hidelinks]{hyperref}
\usepackage[utf8]{inputenc}
\usepackage[small]{caption}
\usepackage{graphicx}
\usepackage{amsmath}
\usepackage{amsthm}
\usepackage{booktabs}
\usepackage{algorithm}
\usepackage{algorithmic}
\usepackage[switch]{lineno}
\usepackage{booktabs}   
\usepackage{graphicx}   

\usepackage{tikz}
\usepackage{comment,verbatim}
\usepackage{amssymb} 
\usepackage{color}
\usepackage{multirow} 

\newcommand{\Emat}[0]{{{\boldsymbol E}}}

\newcommand{\Hmat}[0]{{{\boldsymbol H}}}
\newcommand{\Imat}{{\boldsymbol I}}

\newcommand{\Mmat}[0]{{{\boldsymbol M}}}

\newcommand{\Xmat}{{\boldsymbol X}}
\newcommand{\Ymat}[0]{{{\boldsymbol Y}}}

\newcommand{\pv}[0]{{\boldsymbol{p}}}

\newcommand{\xv}{\boldsymbol{x}}
\newcommand{\yv}{\boldsymbol{y}}

\newcommand{\zv}{\boldsymbol{z}}

\newcommand{\thetav}{\boldsymbol{\theta}}

\title{RobustSCI: Beyond Reconstruction to Restoration for Snapshot Compressive Imaging under Real-World Degradations}

\author{
Hao Wang$^{1,2}$\textsuperscript{\textdagger}\and
Zhankuo Xu$^3$\textsuperscript{\textdagger}\and
Jiong Ni$^2$\and
Xing Liu$^4$\and
Haoyang Liu$^3$\and
Xin Yuan$^{1*}$\\
\affiliations
$^1$Westlake University \and $^2$Xi'an Jiaotong University \\
$^3$Dalian University of Technology \and $^4$Westlake Institute for Optoelectronics \\
\textsuperscript{\textdagger}These authors contributed equally to this work.\\
\emails
*Corresponding author: yuanxin@westlake.edu.cn
}

\begin{document}

\maketitle

\begin{abstract}
Deep learning algorithms for video Snapshot Compressive Imaging (SCI) have achieved great success, yet they predominantly focus on reconstructing from clean measurements. This overlooks a critical real-world challenge: the captured signal itself is often severely degraded by motion blur and low light. Consequently, existing models falter in practical applications. To break this limitation, we pioneer the first study on robust video SCI restoration, shifting the goal from ``reconstruction" to ``restoration"—recovering the underlying pristine scene from a degraded measurement. To facilitate this new task, we first construct a large-scale benchmark by simulating realistic, continuous degradations on the DAVIS 2017 dataset. Second, we propose RobustSCI, a network that enhances a strong encoder-decoder backbone with a novel RobustCFormer block. This block introduces two parallel branches—a multi-scale deblur branch and a frequency enhancement branch—to explicitly disentangle and remove degradations during the recovery process. Furthermore, we introduce RobustSCI-C (RobustSCI-Cascade), which integrates a pre-trained Lightweight Post-processing Deblurring Network to significantly boost restoration performance with minimal overhead. Extensive experiments demonstrate that our methods outperform all SOTA models on the new degraded testbeds, with additional validation on real-world degraded SCI data confirming their practical effectiveness, elevating SCI from merely reconstructing what is captured to restoring what truly happened.
\end{abstract}

\begin{figure}[t]
    \centering
    \makebox[\columnwidth][c]{\includegraphics[width=1.15\columnwidth, keepaspectratio]{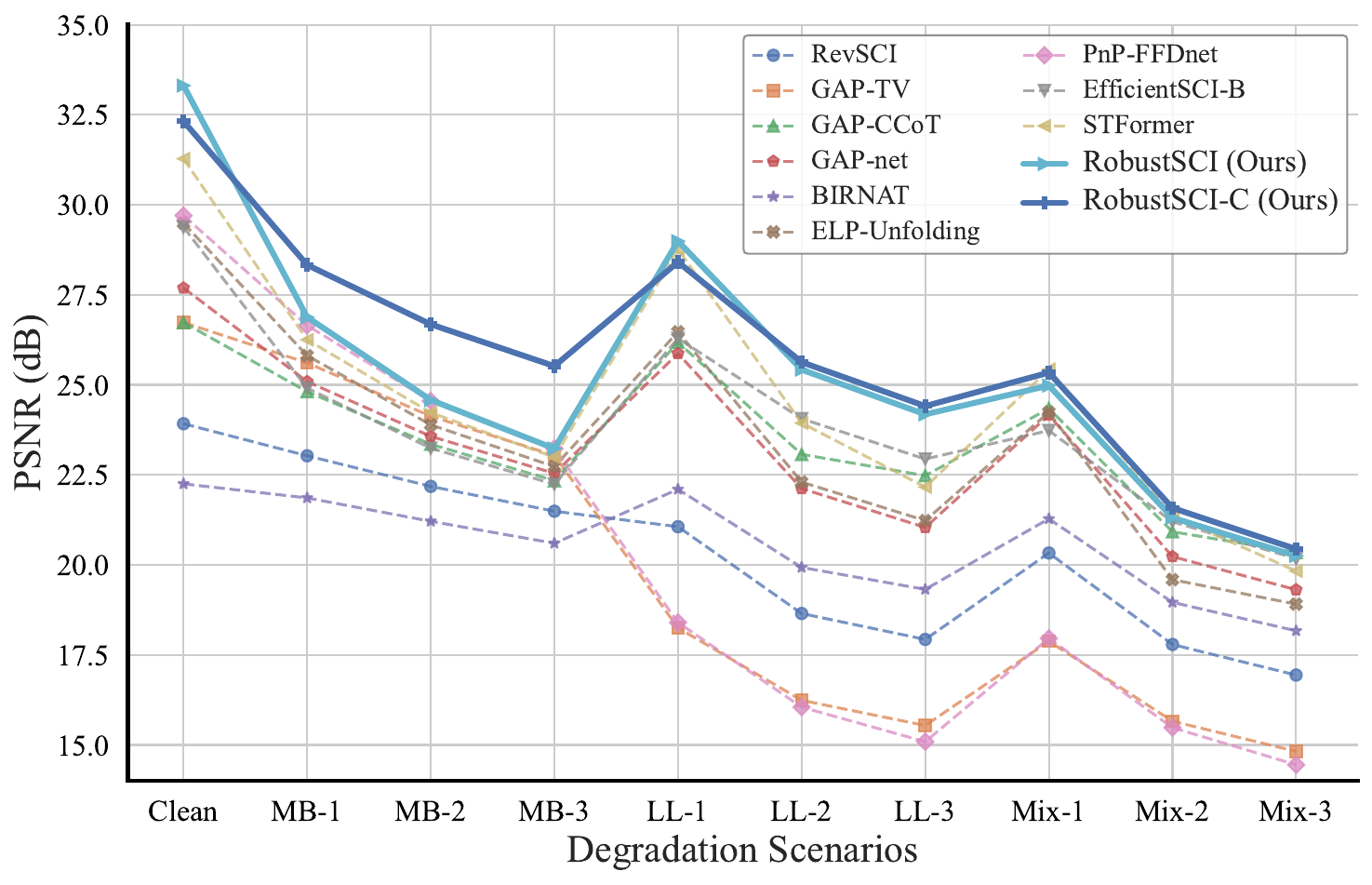}}
    \caption{PSNR performance of our methods against SOTA models across 10 scenarios (clean, 3 motion blur (MB), 3 low-light (LL), 3 mixed degradation levels) on grayscale benchmarks. Our models demonstrate superior robustness as degradation severity increases.}
    \label{fig:psnr_chart}
\end{figure}

\section{Introduction}
Snapshot Compressive Imaging (SCI) presents a novel hardware paradigm for high-speed videography \cite{barbastathis2019use,yuan2016structured}, optically encoding a sequence of video frames into a single 2D measurement to reduce hardware complexity and cost \cite{Llull2013Coded,Yuan2021Snapshot,yuan2014low,candes2006robust}. The core challenge lies in the software decoder, which must solve a highly ill-posed inverse problem to recover the original video frames. While significant progress has been made, existing methods operate under an implicit, idealized assumption: to faithfully reconstruct the optically captured signal.

This assumption overlooks a critical reality: the captured signal is often degraded by motion blur and low-light noise. A perfect reconstruction of this signal yields only a blurry, noisy video, impeding real-world use. Consequently, experimental results show existing models suffer a precipitous performance drop on degraded data. For example, EfficientSCI-B, when trained under the conventional paradigm of pure reconstruction, achieves a PSNR of 36.48 on the clean grayscale test set (detailed in Section~\ref{sec:datasets}); yet its PSNR plummets to 15.90 when evaluated on data degraded with the Mixed-L2 setting (see Table~\ref{tab:degradation_params}).

While prior work addresses specific aspects of SCI robustness \cite{wu2023adaptive,zhang2024event}, restoring pristine scenes from severely degraded measurements remains unsolved. We argue that decoupled two-stage pipelines (reconstruction then restoration) suffer from error accumulation, thus motivating our synergistic, end-to-end paradigm shift from reconstruction to restoration.

To break this bottleneck, we pioneer the first study on robust video SCI with the explicit goal of scene restoration. We redefine the problem: instead of merely reconstructing the degraded input, our objective is to surpass it and recover the underlying, pristine scene. Our key contributions are:
\begin{itemize}
    \item We construct and will release the first large-scale dataset for robust video SCI restoration by simulating realistic degradations (e.g., motion blur and low-light noise) on high-frame-rate videos from DAVIS 2017 \cite{PontTuset2017DAVIS}, followed by a simulation of the SCI measurement process.
    \item We propose the RobustSCI network, which enhances the efficient encoder-decoder architecture of EfficientSCI \cite{wang2023efficientsci} with a novel RobustCFormer block. This new block introduces two parallel restoration branches—a multi-scale deblur branch to explicitly handle motion blur and a frequency enhancement branch to suppress global artifacts—enabling joint reconstruction and restoration.
    \item We further propose the RobustSCI-C (RobustSCI-Cascade) framework, which integrates a pre-trained, lightweight post-processing deblurring network based on NAFNet \cite{Chen2022NAFNet} as a powerful prior, dramatically improving restoration of severe degradations with high efficiency.
\end{itemize}

\begin{figure*}[t]
    \centering
    \includegraphics[width=0.9\textwidth]{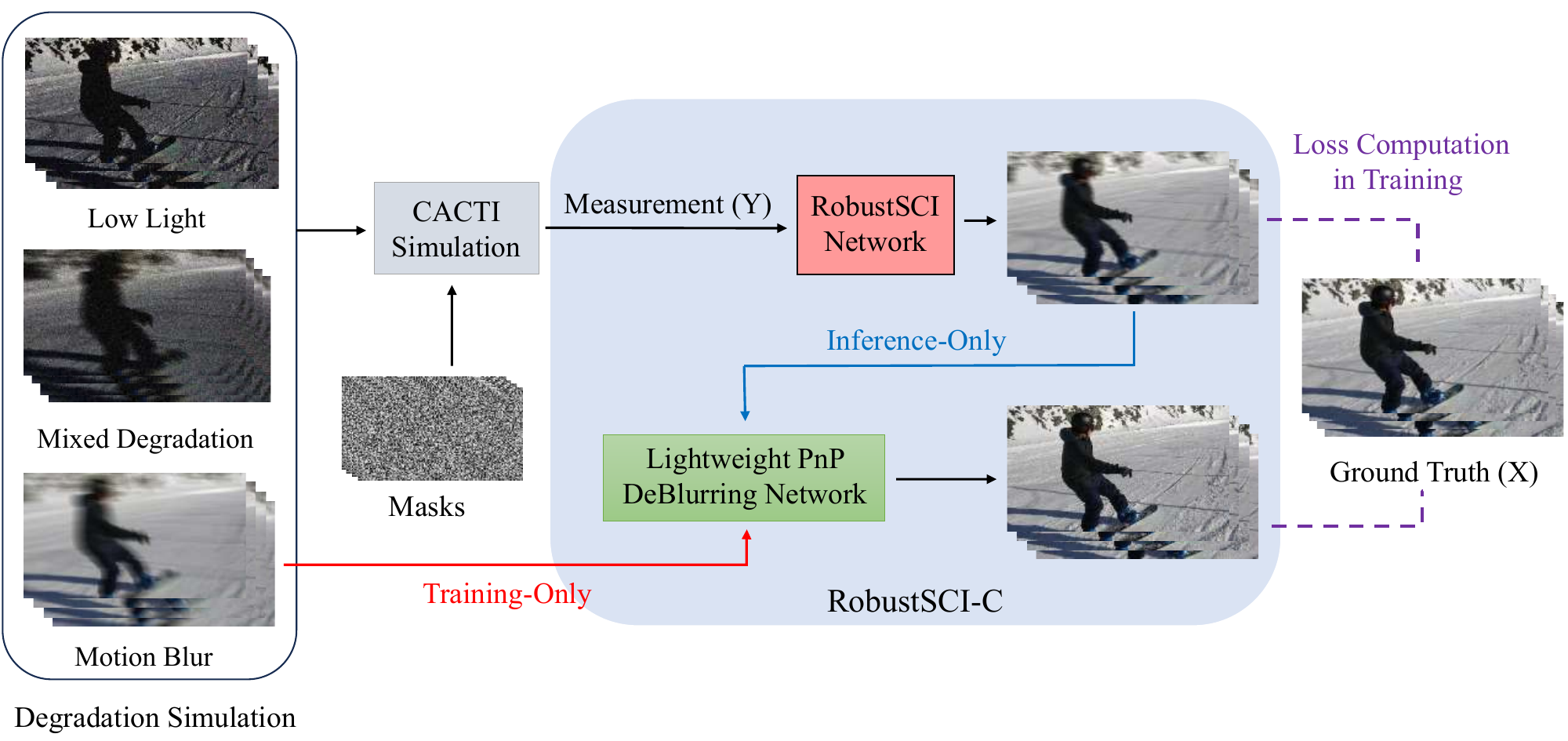}
    \caption{Overview of our training and inference pipeline. Paired data is generated by applying simulated degradations to clean video frames, followed by CACTI simulation to produce degraded measurements ($\Ymat$). RobustSCI is trained end-to-end to recover clean ground truth frames ($\Xmat$), while the Lightweight Post-processing Deblurring Network (trained separately) can be used as a frozen post-processing step during inference.}
    \label{fig:overall_framework}
\end{figure*}

\section{Related Work}
\subsection{Physical Process and Modeling of SCI Systems}
To ground our work in physical reality, it is essential to first understand the optical encoding process of a typical video SCI system (e.g., Coded Aperture Compressive Temporal Imaging, CACTI) and the gap between this physical process and its widely-used simulation in computational studies.

\paragraph{The Physical Process of CACTI.}
As detailed in the survey by Yuan et al. \cite{Yuan2021Snapshot}, a CACTI system optically encodes a high-speed scene into a single 2D measurement to enable high-speed videography with simple hardware. The process relies on a Digital Micromirror Device (DMD) modulator that rapidly displays a sequence of binary coding patterns during the camera’s prolonged exposure. Light from the scene is spatially modulated by these patterns for each temporal frame, and the cumulative modulated light is integrated on the sensor (e.g., CCD/CMOS). This physical process introduces non-idealities—including optical aberrations from lenses, finite DMD switching times/contrast ratios, and sensor noise—that are often ignored in simplified mathematical models.

\paragraph{The Simulated Forward Model.}
In most computational SCI studies (including ours), the forward model is an idealized, discretized approximation of the physical process. Given a sequence of clean video frames $\{\Xmat_k\}_{k=1}^{N_t}$ and pre-defined binary masks $\{\Mmat_k\}_{k=1}^{N_t}$, the measurement $Y$ is computed as:
\begin{equation}
\Ymat = \sum_{k=1}^{N_t} \Xmat_k \odot \Mmat_k + \Emat,
\end{equation}
where $\odot$ denotes element-wise multiplication and $\Emat$ is typically modeled as additive white Gaussian noise (AWGN). This simulation assumes perfect optics, ideal binary masks, and a linear sensor response—simplifications that disconnect it from real-world imaging conditions. Our work bridges this simulation-to-reality gap not by complicating the SCI forward model, but by explicitly injecting realistic degradations $\mathcal{D}$ into clean frames $\Xmat_k$ before they undergo encoding, aligning training data with physical imaging scenarios.

\subsection{SCI Reconstruction Paradigms}
Algorithms for inverting the SCI forward model have evolved from traditional optimization-based methods to deep learning-driven approaches, yet all share a common limitation of assuming pristine input signals. 

Early iterative optimization methods (e.g., GAP-TV \cite{Yuan2016Generalized}, ISTA \cite{bioucas2007new}) use hand-crafted priors (e.g., total variation) for interpretable reconstruction but suffer from slow inference. Deep learning enabled mainstream end-to-end networks \cite{qiao2020deep}, including recurrent (BIRNAT \cite{Cheng2020BIRNAT}), Transformer-based \cite{wang2021metasci,wang2024hierarchical}, and State-Space (MambaSCI \cite{Pan2024MambaSCI}) models. Hybrid approaches like PnP \cite{Yuan2020PnP} (integrating deep denoisers) and Deep Unfolding \cite{Wu2021Dense,Yang2022Ensemble} (combining optimization and deep learning) also emerged. Despite their differences, all these methods invert the idealized forward model (Eq. 1) and ignore real-world signal degradations— a gap RobustSCI addresses.

\subsection{Simulated Degradations}
Since collecting paired real-world degraded/clean SCI data is infeasible, simulating realistic degradations is a cornerstone for training robust restoration models. Our degradation pipeline draws on principled techniques from the deblurring and low-light enhancement communities, aligning with the physical rationale of our RobustSCI design.

\paragraph{Motion Blur Simulation.}
To synthesize motion blur, one common approach involves convolving a sharp image with a blur kernel \cite{levin2009understanding}, though accurately modeling complex motion paths remains a challenge. We instead adopt the more physically-grounded method used by the GoPro dataset \cite{Nah2017GoPro}, which averages consecutive high-speed frames. This technique directly mimics the temporal integration of light inherent in CACTI systems, providing a more realistic simulation of blur from dynamic scenes.

\paragraph{Low-Light Simulation.}
Simulating low-light requires modeling both brightness reduction and noise. While physically accurate models characterize sensor noise as a combination of signal-dependent Poisson and signal-independent Gaussian noise \cite{foi2008practical,chen2018learning}, a widely-used and effective approximation is often employed for training deep models. Following this practical approach \cite{Guo2020ZeroDCE,Feijoo2025DarkIR}, our pipeline first applies a non-linear darkening curve to mimic reduced photon flux, then adds Gaussian noise to approximate the aggregate sensor noise. This provides a tractable yet powerful way to generate diverse low-light conditions.

\section{Methodology}
Our methodology is designed to shift the paradigm of video SCI from conventional reconstruction to robust restoration. We first formalize this problem shift by analyzing the gap between idealized models and physical reality. Based on this analysis, we then detail the architecture of our proposed network, RobustSCI, which is specifically engineered to address these real-world challenges. Our overall strategy is illustrated in Figure~\ref{fig:overall_framework}.

\subsection{Reconstruction to Restoration}
The forward model of video SCI, as established in prior work \cite{Yuan2021Snapshot}, describes how a high-speed video sequence is compressed into a single measurement. In its idealized, vectorized form, this is expressed as:
\begin{equation}
    \yv = \Hmat \xv + \zv,
\end{equation}
where $\xv \in \mathbb{R}^{N_s N_t}$ is the vectorized video sequence, $\Hmat \in \mathbb{R}^{N_s \times N_s N_t}$ is the sensing matrix composed of coding masks, $\yv \in \mathbb{R}^{N_s}$ is the measurement, and $\zv$ represents system noise. The conventional SCI reconstruction paradigm aims to solve for $\xv$ given $\yv$ and $\Hmat$. However, this model makes a critical simplification: it assumes the source signal $\xv$ is pristine.

In a physical imaging system, the signal captured by the optics, $x_c$, has already undergone degradation. Thus, a more physically accurate forward model is:
\begin{equation}
    \yv = \Hmat\xv_c + \zv, \quad \xv_c = \mathcal{D}(\xv_{cl}).
\end{equation}
Here, $x_{cl}$ is the underlying pristine scene, and $\mathcal{D}$ is a complex, non-linear degradation operator that encapsulates cascaded physical effects—including atmospheric disturbances, motion blur from object and camera movement, and photon flux reduction due to low light—instead of arbitrary distortions. The noise $z$ itself is also not merely additive Gaussian; it is a complex mixture of signal-dependent photon shot noise and signal-independent read noise, further complicated by the camera's non-linear response function (CRF).

This reveals a fundamental dichotomy in the problemOur RobustSCI series (RobustSCI/RobustSCI-C) are trained on data with compression ratios definition. The goal of conventional methods is to find an estimate $\hat{x}$ that best approximates $x_{captured}$. In contrast, our goal of scene restoration is to estimate $x_{clean}$. We propose to solve this more challenging problem by training a deep neural network, $\mathcal{G}_{\theta}$, to learn the composite inverse mapping:
\begin{equation}
    \mathcal{G}_{\thetav}(\yv) \approx \xv_{\rm clean}.
\end{equation}
This requires the network not only to invert the sensing matrix $H$, but also to implicitly model and invert the complex degradation operator $\mathcal{D}$. Our network design is therefore explicitly motivated by the need to tackle the core optical effects embedded in $\mathcal{D}$—computationally tractable approximations of real CACTI system artifacts rather than artificial distortions.
Our simulation strategy is designed to model these key physical limitations. Motion blur, which arises from the temporal integration of light during the sensor's exposure, is directly simulated by averaging consecutive high-speed frames. This mimics the amplified blur effect inherent in CACTI systems. To model low-light conditions, which reduce photon flux and introduce a complex mix of shot and read noise, our simulation first applies a darkening curve to represent non-linear intensity compression, followed by the addition of Gaussian noise. This darkening step also implicitly models the degradation of the DMD's modulation contrast under low-light, a critical hardware limitation. Since real-world scenarios often feature these degradations concurrently (e.g., night-time driving), our mixed degradation setup combines these simulated artifacts. This forces the network to learn a robust inversion of intertwined motion and noise effects, thereby tackling real system limitations.

By training on data synthesized with this Optics-inspired degradation model, our network learns to invert not only the compressive encoding $H$ but also the preceding physical degradations $\mathcal{D}$. This aligns with the ultimate goal of computational imaging: to recover the latent scene $x_{\text{clean}}$ from corrupted measurements, pushing SCI from reconstruction towards true scene restoration.

\subsection{Proposed RobustSCI Network}
\begin{figure}[t!]
\centering
\includegraphics[width=\columnwidth]{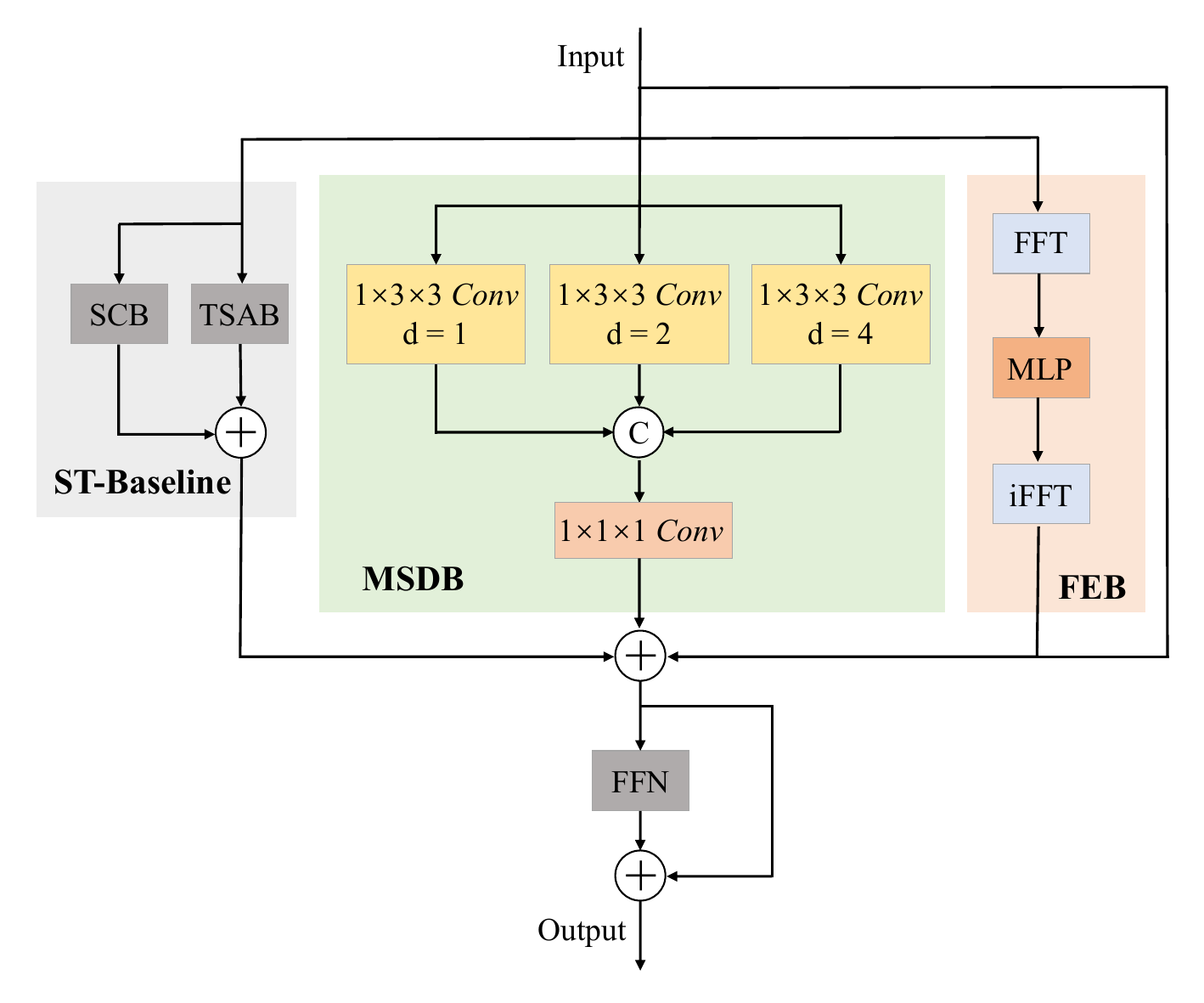} 
\caption{Architecture of the proposed RobustCFormer block. Input features are processed via four parallel branches (ST-Baseline: SCB+TSAB, MSDB, FEB) before fusion. Aggregated features are refined by an FFN with a residual connection.}
\label{fig:robust_cformer}
\end{figure}

To learn the complex mapping in Eq. (3), we propose RobustSCI, which adopts a U-Net-like encoder-decoder backbone with residual connections \cite{he2016deep}. Its core strength lies in the RobustCFormer block (Figure~\ref{fig:robust_cformer}), engineered to disentangle and counteract different degradations via dedicated parallel branches—with feature aggregation and refinement integrated into the block's design.

\paragraph{ST-Baseline Branch.}
Adopting core components from EfficientSCI \cite{wang2023efficientsci}, this branch extracts spatio-temporal features: a Spatial Convolution Branch (SCB) captures local spatial details, while a Temporal Self-Attention Branch (TSAB) models long-range motion patterns, forming the foundation for spatio-temporal data cube reconstruction. The TimesAttention3D module in TSAB computes dynamic frame-wise correlation scores independent of sequence length, enabling our model to handle variable compression ratios (8/16/24) without architectural modifications.

\paragraph{Multi-Scale Deblur Branch (MSDB).}
Motion blur (a spatial degradation from temporal integration) is simulated via kernel-free frame averaging \cite{Nah2017GoPro} (avoiding oversimplified uniform motion assumptions). The MSDB approximates non-uniform blur fields through multi-scale decomposition \cite{Nah2017GoPro,zamir2022restormer}, using three parallel paths with dilation rates $d \in \{1, 2, 4\}$ \cite{luo2020video}: $d=1$ captures fine textures/slow motion (narrow frame windows), while $d=2/4$ expands receptive fields for large-scale motion (wide windows) at no extra cost. Features are concatenated and fused via a $1 \times 1 \times 1$ convolution (learned channel-wise attention) to adaptively select motion scales, outperforming single-scale kernel-based methods for complex motion blur inherent to SCI.

\paragraph{Frequency Enhancement Branch (FEB).}
The FEB addresses global low-light degradations (non-linear intensity compression, wide-spectrum noise) in the frequency domain. Input features are transformed via 2D Real Fast Fourier Transform (RFFT); real/imaginary components of the complex spectrum are concatenated along channels to enable processing by a standard MLP. The MLP learns a dynamic filter to adjust spectral magnitudes/phases—boosting suppressed mid-frequencies for contrast restoration, or attenuating high-frequency noise bands (analogous to learned dynamic equalization). For stability, frequency-domain operations are applied to only half the channels (the other half forms a residual connection); the modified spectrum is transformed back to the spatial domain via inverse RFFT (iRFFT), normalized to mitigate artifacts, and fused with residual channels.

Following processing by the ST-Baseline, MSDB, and FEB branches, all features (including a direct residual connection from the input) are aggregated via element-wise addition. This fused feature map is refined by a Feed-Forward Network (FFN), with a final residual connection producing the block's output—integrating foundational SCI reconstruction with enhancement capabilities tailored for the SCI process.

\subsection{RobustSCI-C Framework}
Observing that a single end-to-end network can struggle to fully eliminate severe motion blur, we propose the RobustSCI-C (RobustSCI-Cascade) framework to explicitly and efficiently address this limitation. This framework operates in two stages. First, the trained RobustSCI network produces an initial, high-quality reconstruction. Second, each frame of this video is independently processed by our Lightweight Post-processing Deblurring Network, which acts as a powerful post-processing prior. This network, based on the NAFNet \cite{Chen2022NAFNet} architecture, is kept frozen during inference and requires no task-specific fine-tuning, making it a highly efficient and effective enhancement module.

\section{Experiments}
\subsection{Experimental Setup}
\paragraph{Datasets.}
\label{sec:datasets}
We adopt the widely-used datasets from BIRNAT \cite{Cheng2020BIRNAT} for both training and evaluation. The training set is derived from the DAVIS 2017 dataset \cite{PontTuset2017DAVIS}. For grayscale evaluation, we use six standard benchmark datasets: Kobe, Traffic, Runner, Drop, Crash, and Aerial. For color evaluation, we use another six benchmarks: Beauty, Bosphorus, Jockey, Runner, ShakeNDry, and Traffic. Crucially, for our robust restoration task, we apply a consistent degradation pipeline to both the training and testing sets. Additionally, we collect real-world degraded SCI data in low-light and motion-blurred scenarios, with a fixed compression ratio (CR) of 24. This real dataset is captured using a custom-built CACTI prototype system, which directly records the degraded compressive measurements of dynamic scenes under natural low-light conditions with inherent motion blur from scene dynamics, providing a practical validation of our methods on real-world imaging conditions.

\paragraph{Degradation Pipeline.}
We design a physics-inspired pipeline to generate realistic degradations for supervised learning. First, all source videos are temporally upsampled 8x via RIFE \cite{huang2022rife} to provide high-frame-rate inputs for accurate motion blur simulation. We then generate paired data $(X_{degraded}, X_{gt})$, where $X_{gt}$ is the central frame of the upsampled sequence, and $X_{degraded}$ is created by applying motion blur and low-light effects:

\begin{itemize}
    \item Motion Blur Simulation: Following the method of the GoPro dataset \cite{Nah2017GoPro}, we average a window of $N$ consecutive high-speed frames centered around the ground truth frame. The parameter $N$, which controls the blur severity, is referred to as Blur ($N$) in our experiments.
    \item Low-Light Simulation: 
    Inspired by Zero-DCE \cite{Guo2020ZeroDCE} and DarkIR \cite{Feijoo2025DarkIR}, we model low-light conditions as a two-stage process encompassing both non-linear brightness reduction and additive sensor noise. For a clean, normalized input image $I$, the final degraded image $I_{degraded}$ at pixel coordinate $p$ is calculated via the model shown in Equation \eqref{eq:low_light_degradation}. Specifically, the first term in the equation models brightness reduction through a quadratic curve controlled by the darkening level parameter $\alpha \in [0, 1]$, while the second term represents additive Gaussian sensor noise (with mean 0 and standard deviation $\sigma$) that is prominent in low-light scenarios. By varying both $\alpha$ and $\sigma$ during data generation, we can synthesize a wide and continuous spectrum of realistic low-light degradations.
\end{itemize}

\begin{equation}
    \label{eq:low_light_degradation}
    \Imat_{degraded}(\pv) = \Imat(\pv) - \alpha \cdot \Imat(\pv) \cdot (1 - \Imat(\pv)) + \mathcal{N}(0, \sigma^2).
\end{equation}

\begin{table}[h]
\centering
\caption{Degradation parameters for the nine test scenarios.}
\label{tab:degradation_params}
\resizebox{0.9\columnwidth}{!}{%
\begin{tabular}{lccc}
\toprule  
\textbf{Scenario} & \textbf{Blur ($N$)} & \textbf{Darkening} & \textbf{Noise ($\sigma$)} \\
\midrule  
MotionBlur-L1 & 7 & - & - \\
MotionBlur-L2 & 11 & - & - \\
MotionBlur-L3 & 15 & - & - \\
\midrule
\addlinespace[0.4em] 
LowLight-L1 & - & 0.6 & 10 \\
LowLight-L2 & - & 0.8 & 25 \\
LowLight-L3 & - & 0.9 & 40 \\
\midrule  
\addlinespace[0.4em] 
Mixed-L1 & 7 & 0.6 & 10 \\
Mixed-L2 & 11 & 0.8 & 25 \\
Mixed-L3 & 15 & 0.9 & 40 \\
\bottomrule  
\end{tabular}
}
\end{table}

\begin{figure*}[t!]
    \centering
    \includegraphics[width=\textwidth]{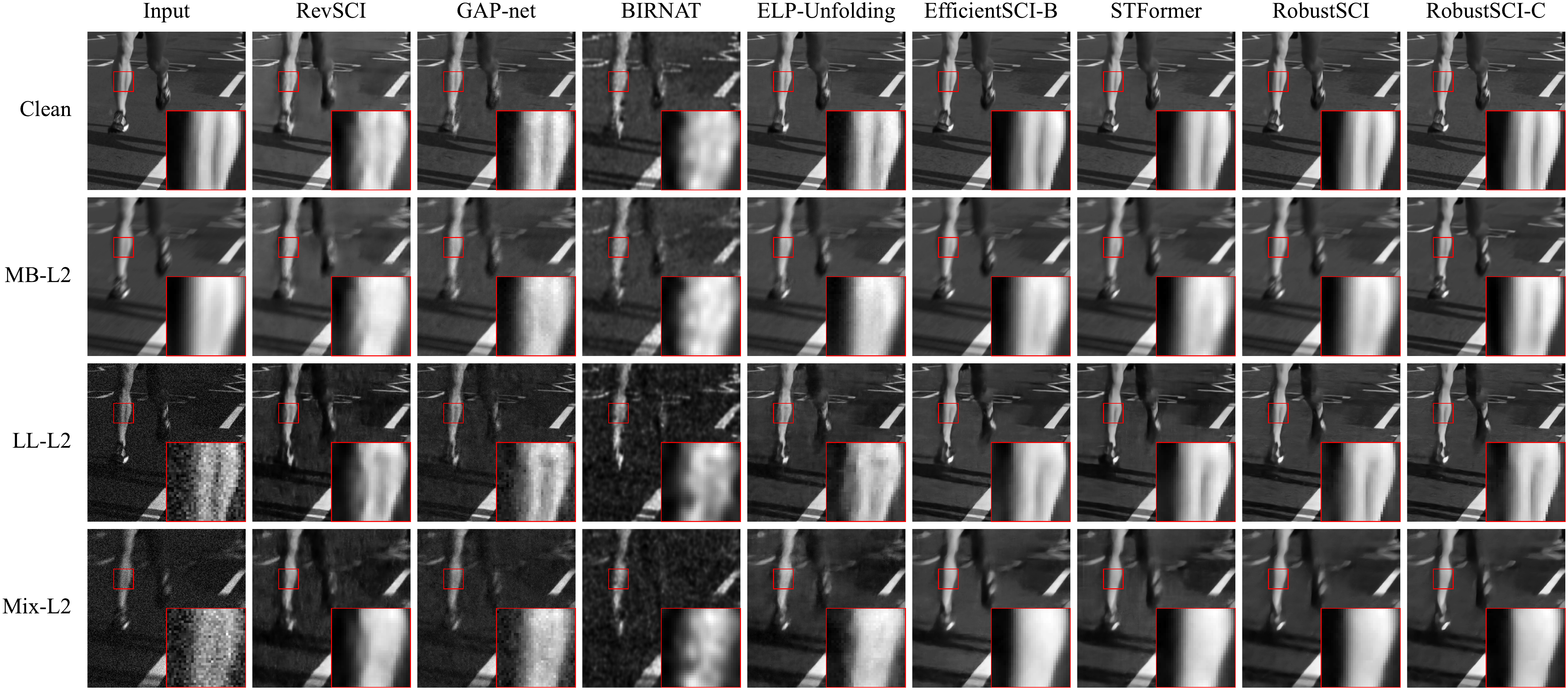}
    \caption{Qualitative comparison on the grayscale runner benchmark under different degradation scenarios. From top to bottom, the rows show results for Clean, MotionBlur-L2 (MB-L2), LowLight-L2 (LL-L2), and Mixed-L2 scenarios. The first column displays the ground truth or the degraded input. Each subsequent column shows the reconstruction from a different model.}
    \label{fig:visual_grayscale_comparison}
\end{figure*}

\begin{table*}[t!]
\centering
\caption{Quantitative comparison (PSNR in dB / SSIM) on degraded grayscale benchmarks. Best results are highlighted.}
\label{tab:grayscale_results}
\scriptsize
\setlength{\tabcolsep}{3.5pt}
\makebox[\textwidth]{
\begin{tabular}{l|cclccccccccccccccccc}
\toprule
\textbf{Method} & \multicolumn{2}{c}{\textbf{Clean}} & \multicolumn{2}{c}{\textbf{MB-1}} & \multicolumn{2}{c}{\textbf{MB-2}} & \multicolumn{2}{c}{\textbf{MB-3}} & \multicolumn{2}{c}{\textbf{LL-1}} & \multicolumn{2}{c}{\textbf{LL-2}} & \multicolumn{2}{c}{\textbf{LL-3}} & \multicolumn{2}{c}{\textbf{Mix-1}} & \multicolumn{2}{c}{\textbf{Mix-2}} & \multicolumn{2}{c}{\textbf{Mix-3}} \\
\midrule
RevSCI & 23.92 & 0.818 & 23.03 & 0.748 & 22.18 & 0.693 & 21.49 & 0.654 & 21.06 & 0.767 & 18.65 & 0.682 & 17.93 & 0.595 & 20.33 & 0.702 & 17.79 & 0.585 & 16.94 & 0.488 \\
GAP-TV & 26.73 & 0.858 & 25.61 & 0.788 & 24.13 & 0.728 & 23.04 & 0.683 & 18.25 & 0.648 & 16.24 & 0.462 & 15.54 & 0.360 & 17.88 & 0.595 & 15.66 & 0.383 & 14.82 & 0.278 \\
GAP-CCoT & 26.73 & 0.895 & 24.81 & 0.801 & 23.35 & 0.728 & 22.34 & 0.677 & 26.18 & 0.848 & 23.07 & 0.746 & 22.48 & 0.626 & 24.34 & 0.764 & 20.92 & 0.611 & 20.35 & 0.493 \\
GAP-net & 27.69 & 0.853 & 25.09 & 0.759 & 23.57 & 0.691 & 22.54 & 0.643 & 25.87 & 0.840 & 22.12 & 0.728 & 21.04 & 0.598 & 24.16 & 0.753 & 20.23 & 0.599 & 19.31 & 0.474 \\
BIRNAT & 22.25 & 0.777 & 21.86 & 0.721 & 21.21 & 0.673 & 20.60 & 0.632 & 22.10 & 0.751 & 19.93 & 0.688 & 19.32 & 0.586 & 21.28 & 0.696 & 18.96 & 0.602 & 18.17 & 0.602 \\
ELP-Unfolding & 29.49 & 0.905 & 25.81 & 0.803 & 23.89 & 0.728 & 22.73 & 0.678 & 26.47 & 0.846 & 22.30 & 0.742 & 21.23 & 0.659 & 24.24 & 0.757 & 19.59 & 0.595 & 18.91 & 0.525 \\
PnP-FFDnet & 29.70 & 0.892 & 26.65 & 0.820 & 24.55 & 0.745 & 23.23 & 0.691 & 18.40 & 0.658 & 16.05 & 0.442 & 15.09 & 0.341 & 17.96 & 0.598 & 15.48 & 0.366 & 14.45 & 0.265 \\
EfficientSCI-B & 29.38 & 0.898 & 24.91 & 0.780 & 23.24 & 0.705 & 22.24 & 0.657 & 26.29 & 0.850 & 24.06 & 0.796 & 22.94 & 0.659 & 23.73 & 0.747 & 21.20 & 0.649 & 20.17 & 0.538 \\
STFormer & 31.28 & 0.940 & 26.25 & 0.819 & 24.23 & 0.740 & 22.99 & 0.686 & 28.76 & 0.884 & 23.94 & 0.803 & 22.16 & \textbf{0.728} & \textbf{25.43} & 0.780 & 21.36 & 0.655 & 19.83 & 0.599 \\
\midrule
RobustSCI (Ours) & \textbf{33.31} & \textbf{0.951} & 26.88 & 0.831 & 24.58 & 0.748 & 23.23 & 0.693 & \textbf{28.99} & 0.898 & 25.42 & 0.823 & 24.18 & 0.727 & 24.98 & 0.794 & 21.31 & 0.673 & 20.25 & 0.613 \\
RobustSCI-C (Ours) & 32.32 & 0.949 & \textbf{28.33} & \textbf{0.872} & \textbf{26.68} & \textbf{0.830} & \textbf{25.52} & \textbf{0.790} & 28.41 & \textbf{0.899} & \textbf{25.63} & \textbf{0.824} & \textbf{24.41} & 0.727 & 25.34 & \textbf{0.810} & \textbf{21.58} & \textbf{0.681} & \textbf{20.44} & \textbf{0.616} \\
\bottomrule
\end{tabular}
}
\end{table*}

For our training set, we generate a diverse and continuous range of degradations by randomly sampling the parameters within their respective ranges (e.g., $N$ from 3 to 17). To simulate dynamic conditions, we ensure a smooth transition of these degradation parameters between consecutive video chunks. For the test set, we define nine fixed degradation scenarios with three severity levels (L1, L2, L3), with specific parameters detailed in Table~\ref{tab:degradation_params}.

\paragraph{Baselines.} We compare our methods against a comprehensive set of SOTA SCI reconstruction algorithms. These include the iterative optimization method GAP-TV \cite{Yuan2016Generalized}; end-to-end deep learning models such as BIRNAT \cite{Cheng2020BIRNAT}, RevSCI \cite{placeholder_revsci}, EfficientSCI-B \cite{wang2023efficientsci}, and STFormer \cite{wang2024hierarchical}; deep unfolding networks including GAP-net \cite{placeholder_gap_net}, GAP-CCoT \cite{placeholder_gap_ccot}, and ELP-Unfolding \cite{Yang2022Ensemble}; and the plug-and-play (PnP) algorithm PnP-FFDnet \cite{Yuan2020PnP}. All learning-based models were retrained on our dataset, while non-learning methods like GAP-TV were evaluated directly.

\paragraph{Metrics.} We use Peak Signal-to-Noise Ratio (PSNR) and Structural Similarity Index (SSIM) \cite{wang2004image} as the primary metrics for quantitative evaluation.

\paragraph{Implementation Details.} 
All experiments are conducted on an NVIDIA RTX 4090 GPU with 24GB memory. Our RobustSCI models are trained on a mixture of data with compression ratios (CR) of 8, 16, and 24, endowing them with flexibility. For a fair comparison, all evaluations reported in this paper, for both our models and the baselines, are conducted on the test set with a fixed CR of 8. We train all models for 150 epochs with a batch size of 4, using the Adam optimizer \cite{kingma2014adam} with an initial learning rate of 1e-4 and cosine annealing learning rate scheduler (decaying to 1e-6 over epochs). Consistent with EfficientSCI \cite{wang2023efficientsci}, we adopt the Mean Squared Error (MSE) loss (PSNR-oriented loss) defined as:
\begin{equation}
\mathcal{L} = \frac{1}{N} \sum_{i=1}^N \|\hat{\Xmat}_i - \Xmat_i\|_2^2,
\end{equation}
where $\hat{\Xmat}_i$ denotes the predicted frames and $\Xmat_i$ is the ground truth. The post-processing deblurring network (based on NAFNet) is trained separately on a dedicated dataset constructed by applying varying levels of motion blur to DAVIS 2017 training frames, with the same optimizer and learning rate settings as the main network.

\begin{table*}[t]
\centering
\caption{Quantitative comparison (PSNR in dB / SSIM) on degraded color benchmarks. Best results are highlighted.}
\label{tab:color_results}
\scriptsize
\setlength{\tabcolsep}{3.5pt}
\makebox[\textwidth]{
\begin{tabular}{l|cclccccccccccccccccc}
\toprule
\textbf{Method} & \multicolumn{2}{c}{\textbf{Clean}} & \multicolumn{2}{c}{\textbf{MB-1}} & \multicolumn{2}{c}{\textbf{MB-2}} & \multicolumn{2}{c}{\textbf{MB-3}} & \multicolumn{2}{c}{\textbf{LL-1}} & \multicolumn{2}{c}{\textbf{LL-2}} & \multicolumn{2}{c}{\textbf{LL-3}} & \multicolumn{2}{c}{\textbf{Mix-1}} & \multicolumn{2}{c}{\textbf{Mix-2}} & \multicolumn{2}{c}{\textbf{Mix-3}} \\
\midrule
EfficientSCI-B & 19.29 & 0.847 & 19.34 & 0.842 & 19.31 & 0.838 & 19.28 & 0.835 & 25.19 & 0.839 & 21.83 & \textbf{0.735} & 21.07 & \textbf{0.626} & 25.22 & 0.835 & 21.50 & \textbf{0.726} & 20.69 & \textbf{0.614} \\
STFormer & 20.48 & 0.763 & 20.54 & 0.761 & 20.48 & 0.757 & 20.40 & 0.753 & 24.73 & 0.803 & 20.58 & 0.717 & 19.56 & 0.604 & 24.51 & 0.799 & 20.34 & 0.710 & 19.29 & 0.594 \\
\midrule
RobustSCI (Ours) & 21.76 & \textbf{0.866} & 22.01 & \textbf{0.860} & 21.77 & \textbf{0.853} & 21.62 & \textbf{0.848} & \textbf{27.93} & \textbf{0.857} & \textbf{23.51} & 0.707 & \textbf{21.90} & 0.581 & \textbf{28.20} & \textbf{0.852} & \textbf{23.30} & 0.698 & \textbf{21.59} & 0.571 \\
RobustSCI-C (Ours) & \textbf{22.64} & 0.845 & \textbf{22.95} & 0.832 & \textbf{22.77} & 0.823 & \textbf{22.64} & 0.818 & 24.12 & 0.823 & 21.82 & 0.715 & 20.75 & 0.607 & 24.15 & 0.814 & 21.66 & 0.705 & 20.55 & 0.595 \\
\bottomrule
\end{tabular}
}
\end{table*}

\begin{figure}[h!]
    \centering
    \includegraphics[width=\columnwidth]{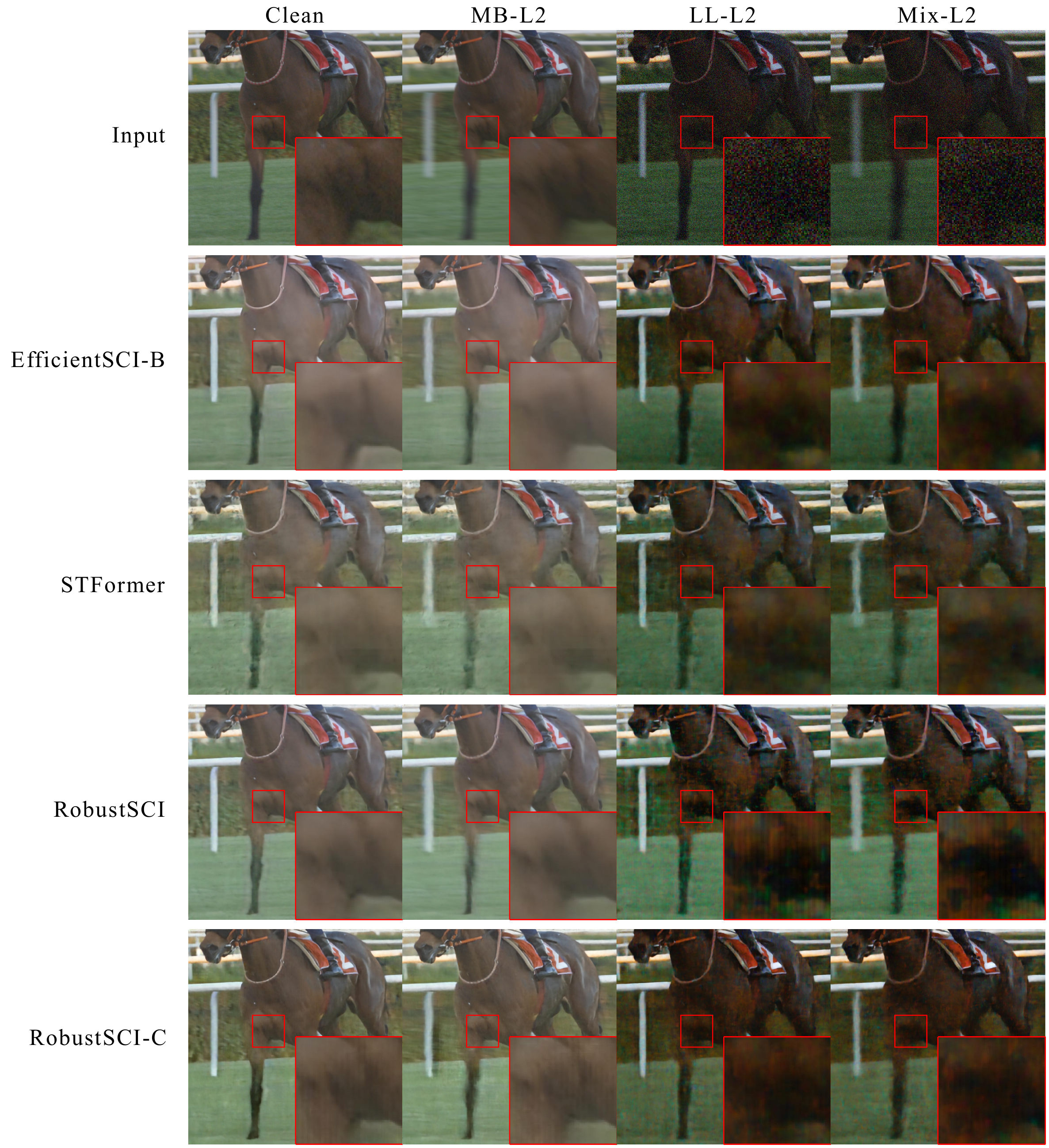}
    \caption{Qualitative comparison on the color Jockey benchmark. From top to bottom, the rows show the input and the results from different models. From left to right, the columns show different degradation scenarios at the L2 level.}
    \label{fig:visual_color_comparison}
\end{figure}

\subsection{Results on Grayscale Videos}
We first present the comprehensive quantitative and qualitative results on the degraded grayscale benchmarks. 

Quantitative comparisons with mainstream baselines in Table~\ref{tab:grayscale_results} and Figure~\ref{fig:psnr_chart} demonstrate the superiority of our methods. RobustSCI and RobustSCI-C consistently outperform others across nearly all ten test conditions, from clean data to severe degradations, with the performance gap widening as degradation intensifies. This trend holds across the board; RobustSCI and RobustSCI-C dominate the quantitative comparisons, claiming nearly all the top results in Table~\ref{tab:grayscale_results} and thus verifying the robustness of our restoration-oriented design.
Figure~\ref{fig:visual_grayscale_comparison} validates these findings qualitatively.  Moreover, RobustSCI-C delivers clearer edges and sharper details in motion blur cases, confirming the effectiveness of our lightweight post-processing deblurring network.

\subsection{Results on Color Videos}
As shown in Table~\ref{tab:color_results}, our methods maintain their significant lead on the degraded color video benchmarks. We compare against strong baselines, including EfficientSCI-B and STFormer, which were retrained on our robust training dataset. The RobustSCI-C framework, in particular, demonstrates exceptional performance in restoring both image structure and color fidelity. A qualitative comparison on the Jockey dataset is provided in Figure~\ref{fig:visual_color_comparison}. The visual results confirm the quantitative findings, highlighting the severe compromises in the outputs of baseline models. In contrast, our method produces visually pleasing videos with sharp details, natural colors, and minimal artifacts, whereas other methods suffer from color shifts, residual noise, and blur. This underscores the effectiveness of our proposed architecture and training strategy for the challenging task of robust color video SCI restoration.

\subsection{Ablation Study}
To validate the effectiveness and efficiency of our proposed components, we conduct an ablation study. We start with a baseline (EfficientSCI-B retrained on our data) and incrementally add our proposed branches. We evaluate performance by averaging PSNR and SSIM scores across all ten grayscale test conditions, and also report the computational complexity in GFLOPs. As shown in Table~\ref{tab:ablation}, adding either the MSDB or the FEB individually brings a notable performance gain over the baseline with only a modest increase in parameters and GFLOPs. Integrating both branches to form our full RobustSCI model yields further improvement, demonstrating their complementary nature while keeping the computational cost manageable. Finally, the Lightweight Post-processing Deblurring Network provides the largest performance boost but also substantially increases the total GFLOPs, highlighting the trade-off between peak performance and computational efficiency.

\begin{table}[h!]
\centering
\caption{Ablation study of our proposed components. Performance and GFLOPs are averaged over all 10 grayscale test conditions.}
\label{tab:ablation}
\adjustbox{max width=\linewidth, center}{
\begin{tabular}{lcccc}
\toprule
\textbf{Method} & \textbf{Params (M)} & \textbf{FLOPs (G)} & \textbf{PSNR} $\uparrow$ & \textbf{SSIM} $\uparrow$ \\
\midrule
Baseline & 8.82 & 1423.6 & 23.82 & 0.728 \\
Baseline + MSDB & 9.682 & 1533.9 & 24.92 & 0.76 \\
Baseline + FEB & 9.88 & 1493.4 & 24.54 & 0.754 \\
RobustSCI  & 10.74 & 1603.7 & 25.31 & 0.775 \\
RobustSCI-C & 18.04 & 1662.4 & 25.87 & 0.800 \\
\bottomrule
\end{tabular}
}
\end{table}

\subsection{Results on Real-World Degraded SCI Data}
\begin{figure}[t!]
\centering
\includegraphics[width=\columnwidth]{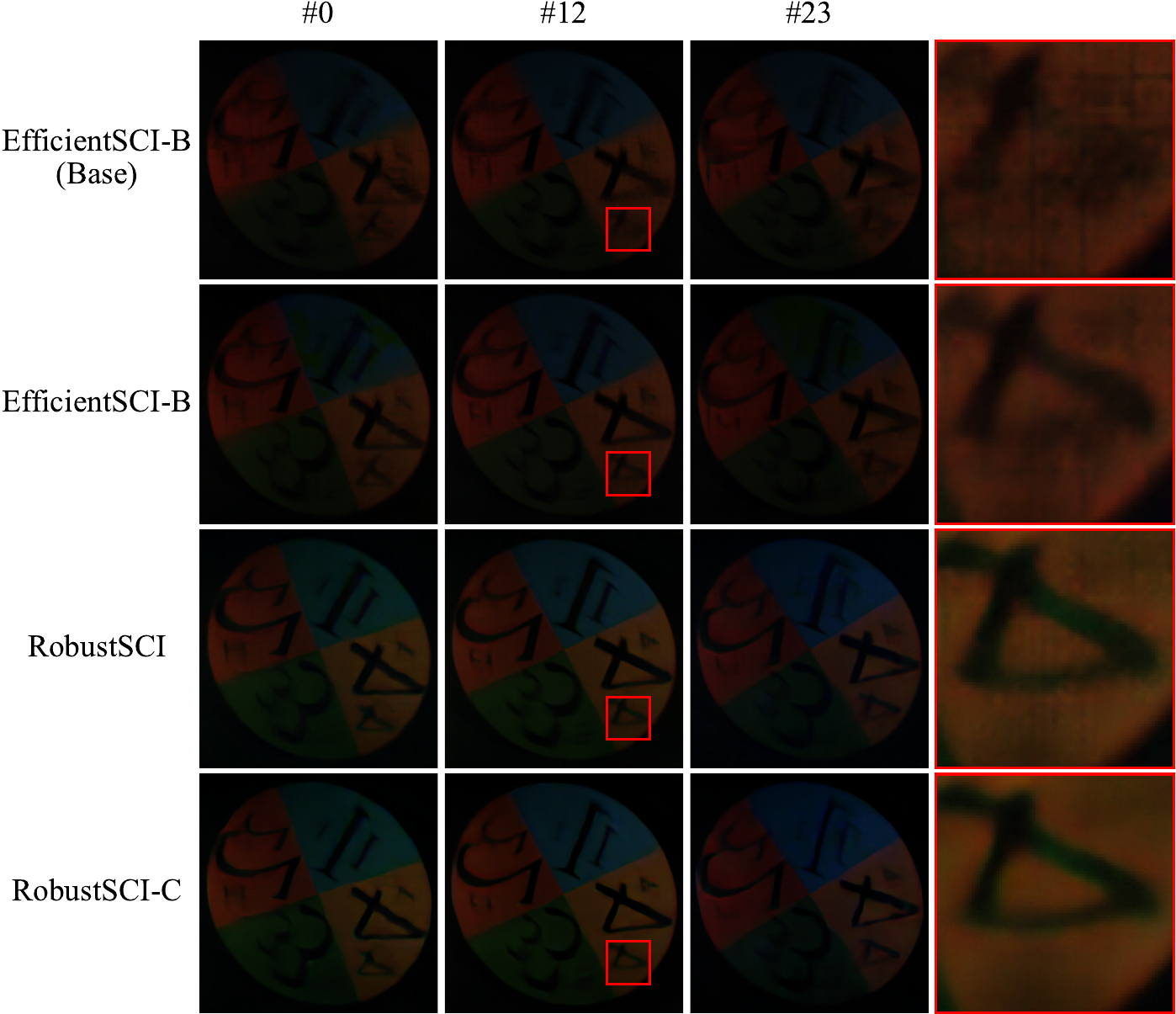}
\caption{Qualitative comparison on real-world degraded SCI data. From top to bottom: EfficientSCI-B (Base, trained under the conventional degradation-agnostic paradigm), EfficientSCI-B, RobustSCI, and RobustSCI-C. The rightmost column shows a low-light enhanced version of the boxed region for better visualization.}
\label{fig:real_world_results}
\end{figure}

To validate practical effectiveness, we evaluate RobustSCI and RobustSCI-C on real-world degraded SCI data (CR=24) captured by an actual optical CACTI system (low-light + motion blur), and compare them with two versions of EfficientSCI-B (EfficientSCI-B (Base), trained under the conventional degradation-agnostic reconstruction paradigm, and the standard EfficientSCI-B trained under our experimental settings). Since no ground-truth frames are available for real data, we only provide qualitative comparisons, and as shown in Figure~\ref{fig:real_world_results}, our methods outperform both versions of EfficientSCI-B in suppressing noise/blur and recovering clear details. These results confirm the generalization of our methods to real SCI scenarios.

\section{Conclusion}
This paper pioneers the shift in video SCI from reconstruction to restoration, addressing the problem of common real-world degradations like motion blur and low light. We introduce a large-scale benchmark for this task and propose RobustSCI, a novel network architecture designed for joint reconstruction and restoration, with its RobustCFormer block enabling targeted disentanglement of spatial-temporal blur and frequency-domain noise. We further boost performance with RobustSCI-C (RobustSCI-Cascade), an effective and efficient post-processing framework requiring no retraining. Extensive experiments demonstrate that our methods establish a new state-of-the-art on degraded data, paving the way for practical SCI applications in challenging environments.

\appendix

\bibliographystyle{named}
\bibliography{ijcai26} 

\end{document}